\documentclass[letterpaper, 10 pt, conference]{ieeeconf}  

\IEEEoverridecommandlockouts                              

\usepackage{multirow}
\usepackage{amsmath}  
\usepackage{amssymb}  
\usepackage[colorlinks,urlcolor=blue,linkcolor=blue,citecolor=blue]{hyperref}
\usepackage{color,array}
\usepackage{xcolor}
\usepackage{graphicx}
\usepackage{setspace}  
\usepackage{subcaption}
\usepackage{caption}
\DeclareCaptionLabelSeparator{periodspace}{.\quad}
\usepackage{graphics} 
\usepackage{epsfig} 
\usepackage{times} 
\usepackage{amsmath} 
\usepackage{amssymb}  
\usepackage[linesnumbered,ruled,vlined]{algorithm2e}

\usepackage{multirow} 
\usepackage{booktabs} 
\usepackage{gensymb}  
\usepackage{bm}  
\usepackage{underscore}  

\usepackage{amsmath} 
\usepackage{amssymb}  

\usepackage{threeparttable}

\usepackage{bbding} 
\usepackage{cite}
\usepackage{hyperref}
\hypersetup{
colorlinks  = true,
citecolor = blue,
linkcolor = blue,   
urlcolor = blue
}

\author{Hanjing Ye$^{1}$, Jieting Zhao$^{1}$, Yaling Pan$^{2}$, Weinan Chen$^{2}$, Li He$^{1}$ and Hong Zhang$^{1*}$
\thanks{*corresponding author (hzhang@sustech.edu.cn).}
\thanks{$^{1}$Hanjing Ye, Jieting Zhao, Li He and Hong Zhang are with Shenzhen Key Laboratory of Robotics and Computer Vision, Southern University of Science and Technology (SUSTech), and the Department of Electronic and Electrical Engineering, SUSTech.}%
\thanks{$^{2}$Yaling Pan and Weinan Chen are with the Biomimetic and Intelligent Robotics Lab, Guangdong University of Technology.}%
\thanks{This work was supported by the Pearl River Talent Recruitment Program under Grant No.2019QN01X761 and the National Natural Science Foundation of China (62103179).}%
\thanks{Source Code: \url{https://github.com/MedlarTea/Mono-RPF}.}%
}

\begin{document}
\title{\LARGE \bf
Robot Person Following Under Partial Occlusion
}
\thispagestyle{empty}
\pagestyle{empty}

\maketitle

\begin{abstract}
        Robot person following (RPF) is a capability that supports many useful human-robot-interaction (HRI) applications. However, existing solutions to person following often assume full observation of the tracked person. As a consequence, they cannot track the person reliably under partial occlusion where the assumption of full observation is not satisfied. In this paper, we focus on the problem of robot person following under partial occlusion caused by a limited field of view of a monocular camera. Based on the key insight that it is possible to locate the target person when one or more of his/her joints are visible, we propose a method in which each visible joint contributes a location estimate of the followed person. Experiments on a public person-following dataset show that, even under partial occlusion, the proposed method can still locate the person more reliably than the existing SOTA methods. As well, the application of our method is demonstrated in real experiments on a mobile robot.
\end{abstract}

\section{INTRODUCTION}
Robot person following (RPF) \cite{islam2019person} is a capability that supports many useful HRI\cite{goodrich2008human} applications. Often, the person being followed can become partially occluded in various situations due to, for example, other objects or people in the robot working environment. Therefore, the ability of following a person under partial occlusion is essential.

RPF can be achieved with a distance measurement sensor such as a LiDAR and or an RGB-D camera. Methods in\cite{leigh2015person, sung2015hierarchical, yuan2018laser, wang2017real, koide2016identification, linder2016multi}, for example, firstly track multiple people with the help of a distance measurement sensor. Once a target person is selected and tracked in the field of view of the robot sensor, the person can be followed on the basis of the tracked location. Such solutions, however, can be expensive due to the high cost of a distance measurement sensor. In addition, distance sensors lack textural information, and this prevents them from resolving data association effectively. 

\begin{figure}[t]
        \centering
        \begin{subfigure}[]{0.43\linewidth}
                \centering
                \includegraphics[width=\linewidth]{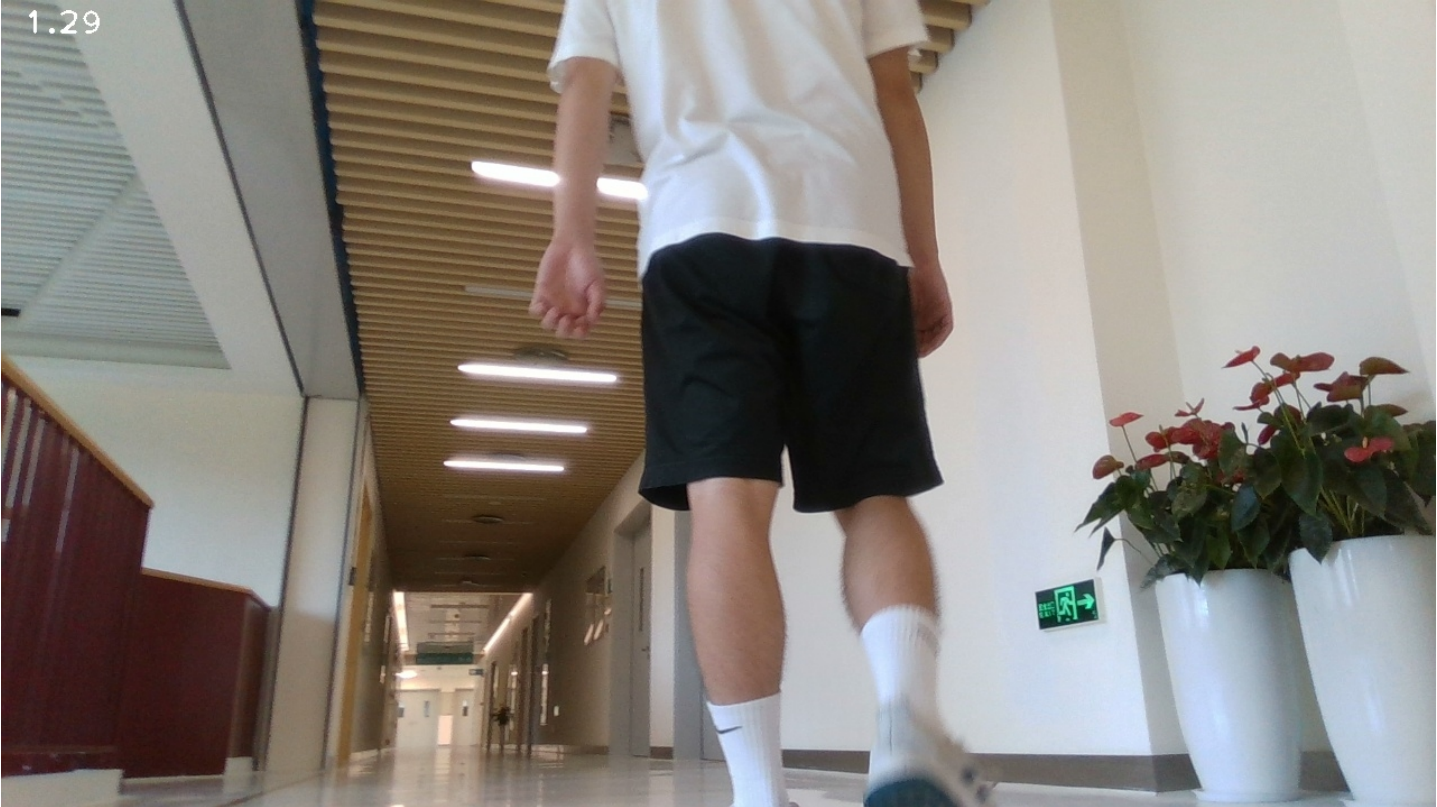}
                \caption{Partial occlusion}
                \label{introduction-a}
        \end{subfigure}%
        \hspace{0.03\linewidth}
        \begin{subfigure}[]{0.43\linewidth}
                \centering
                \includegraphics[width=\linewidth]{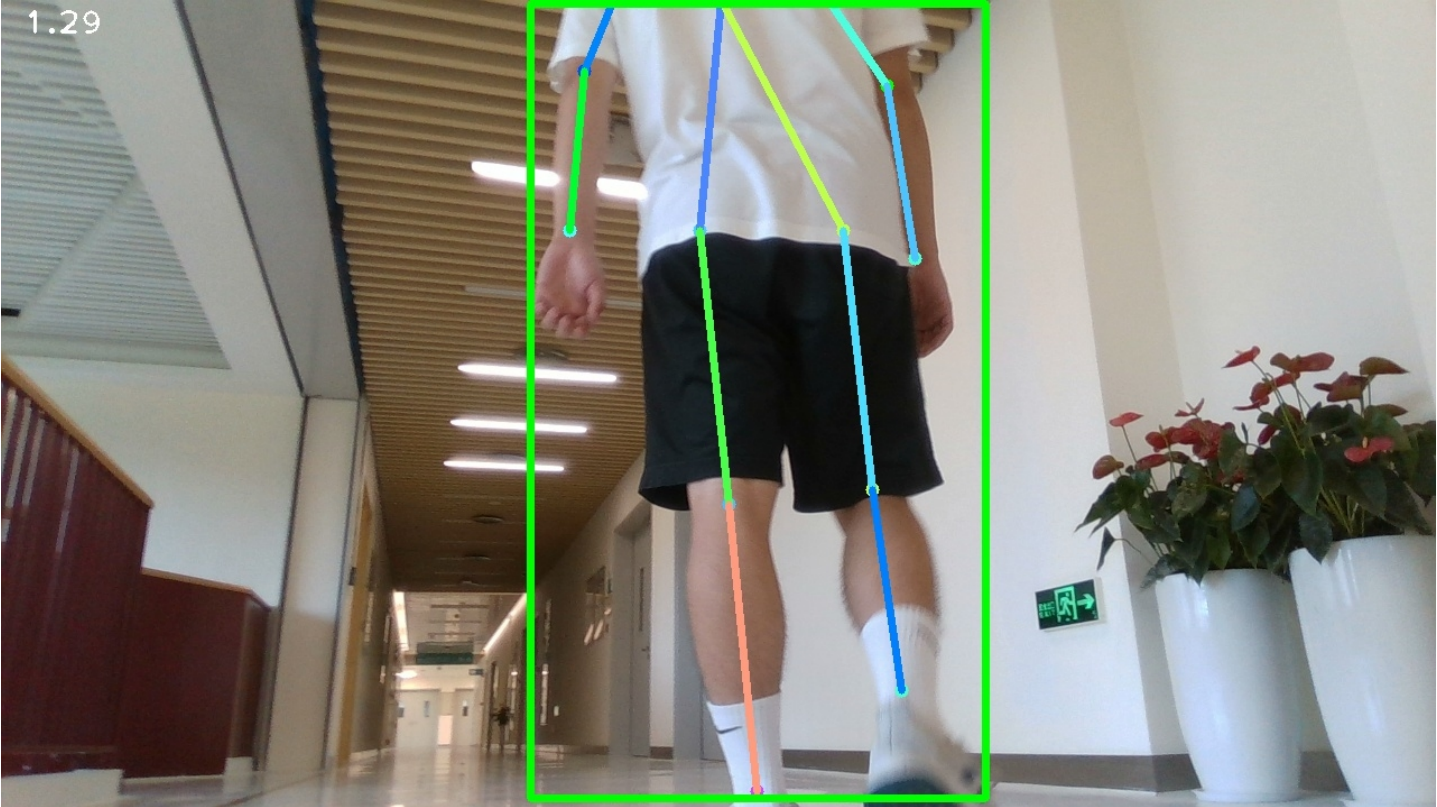}
                \caption{Visible observations}
                \label{introduction-b}
        \end{subfigure}%
        \vspace{0.02\linewidth}
        \begin{subfigure}[]{0.9\linewidth}
                \centering
                \includegraphics[width=\linewidth, height=0.5\linewidth]{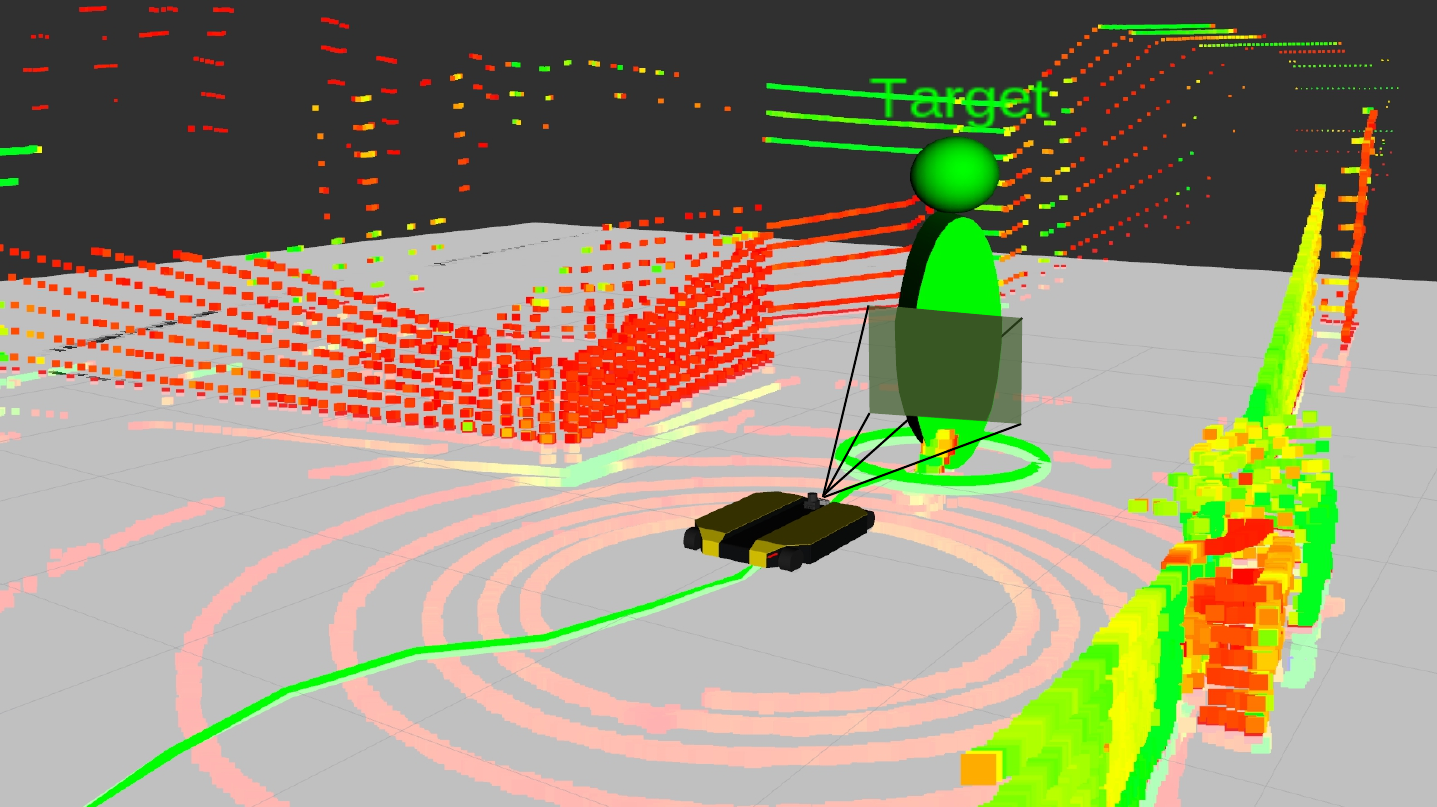}
                \caption{Target person location and following}
        \end{subfigure}%
\caption{(a) An example of partial occlusion caused by a limited FoV of a camera while performing person following. Existing methods often fail to locate the person in such challenging scenarios. (b) The proposed method is based on partial observations, including the bounding box and 2D joints. (c) Person following under partial occlusion (the pointcloud is used as a reference for visualization only, and the body parts of the detected person (target) are marked in green solids in front of the robot).}
\label{introduction}
\vspace*{-0.30in}
\end{figure}

Alternatively, one can resort to vision to solve the problem of person following. \cite{zhang2019vision} uses a vision-based single object tracker (SOT) \cite{zheng2021improving, li2019siamrpn++, danelljan2019atom} to track the person in the image space, and relies on visual servo to follow the person.
Also using vision, \cite{koide2020monocular} proposes a method under the assumption that the neck of the person is always visible. Conceptually, it can be considered as a \textit{single-joint-based} method. However, with a camera of limited field-of-view (FoV), neck visibility cannot always be guaranteed even though the followed person may still be partially visible, as shown in the example in Fig.~\ref{introduction}(a). 
In addition, many \textit{deep-learning-based} methods have been developed for estimating a person's location through monocular depth prediction\cite{monodepth2}, monocular 3D bounding-box detection\cite{park2021dd3d}, etc. These methods, however, are known to experience poor generalization in terms of adapting to the case of partial observation. 

One prior topic of research related to our study is 2D human joint detection\cite{openPose}\hspace{1pt}\cite{alphaPose}, which has been well developed in recent years. It has been demonstrated that human joints can be detected reliably even under partial observation as depicted in Fig.~\ref{introduction}(b). In this work, We exploit these well-detected joints of a partially occluded person in person location estimation. This idea of estimating the 3D pose of an object or a person from partial observation is not new\cite{cheng2019occlusion, bogo2016keep, nguyen2022templates}. To design a solution, one can first build a prior model describing the physical attributes of as well as constraints between the parts of the whole object/person, and this model can be created in the form of an implicit representation by a neural network\cite{cheng2019occlusion}, a parametric model like SMPL (skinned multi-person linear model)\cite{bogo2016keep}, or a CAD model\cite{nguyen2022templates}. Then given a partial observation, the 3D person/object poses can be inferred from the prior model.

To make use of the above idea in our study of RPF, without loss of generality, we build a prior model consisting of the heights of a person's joints relative to the ground plane. With the assumption of a standing person on the ground plane and a calibrated camera at a known height to the ground, the person's location with respect to the camera can be inferred from any joint detection given the prior model\cite{hoiem2008putting}.
In fact, each visible joint contributes a location estimate and the concurrent observations of multiple joints can be fully utilized through any one of the many existing sensor fusion techniques such as the Kalman filter. 
Furthermore, to handle a crowded scene with multiple people, we perform people tracking and target re-identification by using the bounding boxes of detected people and an appearance model of the followed person respectively.
Lastly, a robot motion control module is implemented to form a complete RPF solution, which we refer to as \textit{visible-joint-based} RPF method. In experiments, our method can follow a target person reliably even under partial occlusion, as shown in Fig.~\ref{introduction}(c).

\section{RELATED WORK} \label{sec:related-works}

\subsection{Monocular Person Tracking in Robot Person Following} \label{sec:related-works-a}
A person following robot usually involves four modules: a detection module, a tracking module, an identification module and a robot motion control module. The detection module detects the tracked person in the image. The tracking module locates the person in terms of a 3D pose with respect to the robot. The identification module usually maintains an appearance model useful for re-identification in case of a lost target. Given the person's estimated location, the robot motion control module computes a motion command for the robot to maintain a desired following position with respect to the target person.

Many existing works in people tracking \cite{leigh2015person,sung2015hierarchical,yuan2018laser,wang2017real,koide2016identification,linder2016multi} use distance measurement sensors, which can be expensive and have difficulty in dealing with cluttered indoor environments due to the lack of textural information that is critical for data association. Some works are based on monocular vision. \cite{zhang2019vision} tracks a person in the image space, and achieves person following with visual servo by keeping the person in the center of the image plane . 
Image-based tracking is convenient to implement but ineffective compared to the position-based tracking in 3D as the person and the robot move physically in the Cartesian space, rather than the 2D image space\cite{choi2010multiple}.

For estimating a person's location in the robot coordinate frame, inspired by well-known techniques in video surveillance\cite{hoiem2008putting}\hspace{1pt}\cite{choi2010multiple}, \cite{koide2020monocular} proposes a \textit{single-joint-based} method to locate the person by assuming that the neck of the person is always visible.
However, this assumption cannot always be satisfied. Given the fact that under partial occlusion, one or more joints (not necessarily the neck) are still visible, it should be possible to estimate a person's location from the observed joints. Therefore, we propose to use the observable joints to track the person instead of a specific joint to address the challenge of partial occlusion.

\subsection{2D Human Bounding-box Detection and Joint Detection}
To develop the detection module of our RPF method, there are two related topics: 2D human bounding-box detection\cite{liu2020deep} and joint detection\cite{toshev2014deeppose}. 2D human bounding-box detection\cite{toshev2014deeppose} seeks to locate people as bounding boxes in an image, and the methods can be divided into two families: two-stage\cite{girshick2015fast} and one-stage\cite{ge2021yolox}. A two-stage\cite{girshick2015fast} method usually first generates category-independent proposals and then utilizes category-specific classifiers to label the proposals, while a one-stage method\cite{ge2021yolox} directly generates labeled bounding boxes without any proposal generation. In this paper, we use a one-stage method (YOLOX\cite{ge2021yolox}) to detect people for its fast inference and stable performance.

2D human joint detection\cite{toshev2014deeppose} has also been well developed in recent years. It aims at localizing human joints (keypoints) in an image. There are also two families of methods: top-down\cite{alphaPose}\hspace{1pt}\cite{sun2019deep} and bottom-up\cite{newell2017associative}\hspace{1pt}\cite{CAO}. A top-down method first detects bounding boxes, and then localizes 2D joints within the boxes, whereas a bottom-up method detects joints first and then groups them into full bodies. We adopt a top-down method (AlphaPose\cite{alphaPose}) in the tracking module of our \textit{visible-joint-based} RPF system, due to its good performance even under partial occlusion.

\subsection{3D Person Location Estimation} \label{3d-person-location}
A RPF system must contain a tracking module for person location estimation, and the current leading methods in solving this problem are mostly deep-learning-based. Such methods employ a deep neural network to infer the location of a person from the observed image in an end-to-end fashion.
MonoLoc\cite{monoloc} first uses a neural network to detect joints of a person in an image, and then utilizes these estimated 2D joint positions to locate the person in 3D by a multi-task neural network.
EPro-PnP\cite{epropnp} describes the pose of a person in the form of a 3D bounding box by integrating learnable 2D-3D correspondences.
RootNet\cite{mono3dpose} develops a top-down pose estimation solution that computes the 3D poses of multiple people with respect to the camera coordinate frame.

\begin{figure*}[t]
        \centering
        \includegraphics[width=0.9\linewidth]{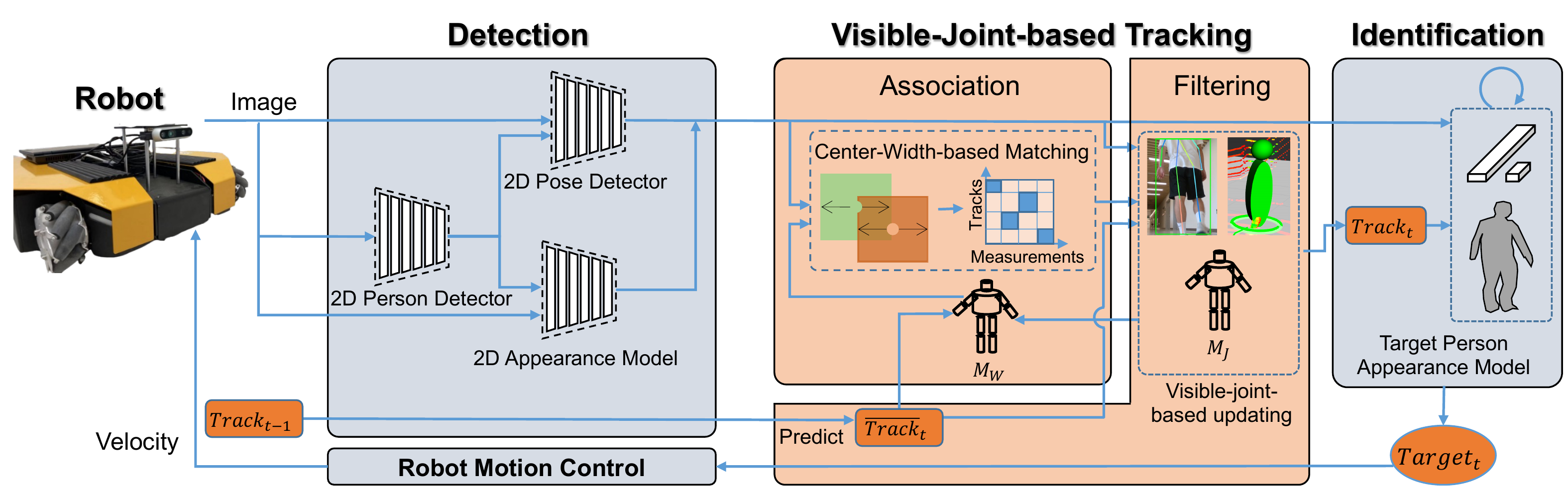}
        \caption{Our proposed \textit{visible-joint-based} RPF system is composed of a Detection module(Sec.~\ref{detection}) including 2D person detector, 2D pose detector and appearance extraction model, a tracking module with track initialization (Sec.~\ref{initialization}), filtering (Sec.~\ref{filtering}), and data association (Sec.~\ref{data-association}), an identification module (not our focus in this paper) and a robot motion control module. $\mathcal{M}_J$ is the prior global model consisting of heights of joints and $M_W$ represents the person's real width. This system can locate, track, and follow the target person even under partial occlusion.}
        \label{method-framework}
        \vspace*{-0.25in}
\end{figure*}

All the methods mentioned above are entirely based on deep learning, although MonoLoc\cite{monoloc} consists of two separate networks, one for joint detection and the other person location, in a way that is similar to our solution. In addition, their training datasets usually involve a full observation of the objects of interest. However, in the person following scenario, a partial observation of the person often occurs, a situation that these methods cannot deal with effectively. In our work, we adopt a hybrid approach with a detection module and a tracking module where a learning-based detector\cite{alphaPose} provides 2D information about the joints of the target person in an image to a subsequent model-based 3D person location tracker. The main advantages of this approach are: 1) 2D joint detectors, compared to the \textit{deep-learning-based} methods, are shown to be more robust with respect to partial occlusion and 2) our proposed model-based location tracker is able to utilize a prior model of the target person that easily computes the person's location analytically from these well-detected 2D joints.

\section{METHODOLOGY} \label{sec:methodology}
We introduce a novel \textit{visible-joint-based} method by utilizing any visible joints to track the target person instead of a specific joint. Together with an identification module and a robot motion control module (which are not the focus in this paper), a complete RPF system is formed. 
The general overview of our system is shown in Fig.~\ref{method-framework}. 

Typically, our tracking method follows the general process of the Kalman filter.
In model construction (Sec.~\ref{initialization}), from detections of all joints in the first frame, we first build a prior model $\mathcal{M}_{J}$ consisting of the heights of the joints with respect to the ground plane. Then upon the detection of a target person, the target person's location can be initialized with the help of this prior model. 
In addition, in the filtering stage (Sec.~\ref{filtering}), tracks are updated with associated joint measurements where each visible joint contributes a location estimate, also with the help of the prior model. 
In the data association (Sec.~\ref{data-association}), we associate tracks and joint measurements in the bounding-box-like space due to the stability of the bounding-box detector. In case of target loss, our RPF system uses the identification module to re-identify a target person.
Lastly, with the robot motion control module, person following is performed in the robot coordinate frame based on the estimated location of the target person.

As in existing works\cite{koide2020monocular}\hspace{1pt}\cite{choi2010multiple}, the x-y plane of the robot coordinate frame is parallel to the ground plane. In addition, we use a calibrated camera whose origin has a known offset with respect to the robot coordinate frame and whose orientation is identical to the robot coordinate frame except for a known positive tilt angle (see Fig.~\ref{method-raycasting}).

\subsection{Detection Module} \label{detection}
First, we detect people to generate their bounding boxes and 2D joint positions in the image plane. We use the bounding boxes and joint positions for tracking the target person, and we use the appearance features within the bounding boxes for target identification. Specifically, we use YOLOX\cite{ge2021yolox} for bounding-box detection and AlphaPose\cite{alphaPose} for joint detection. We exploit all detected joints of a partially occluded person in his location estimation.

In each image, we define as our observation of a person--detected bounding box and joint positions including shoulder, hip, knee and ankle in the image plane. They are represented as $\mathcal{D}=\{\mathcal{B},\mathcal{P}\}$, where $\mathcal{B}=\{[u,v],w,h\}$ defines the center, the width and the height of the bounding box, and $\mathcal{P}$ is the set of visible joints--a subset of $\{ \mathbf{p}_{neck},\mathbf{p}_{hip},\mathbf{p}_{knee},\mathbf{p}_{ankle}\}$ where $\mathbf{p}_i\in \mathbb{R}^2$, describes the pixel coordinates of a visible joint. Note that in our study, we describe each of the three pairs of hip, knee, and ankle joints by a single 2D point $\mathbf{p}_i$ in the image plane. The horizontal coordinate of $\mathbf{p}_i$ is that of the bounding box of the detected person; the vertical coordinate of $\mathbf{p}_i$ is the average height of the both joints or that of a single joint, in the pair, depending upon if one or two joints of the pair are visible.

\subsection{Prior Model Construction and Track Initialization}  \label{initialization}
\begin{figure}[t]
        \centering
        \includegraphics[width=0.9\linewidth]{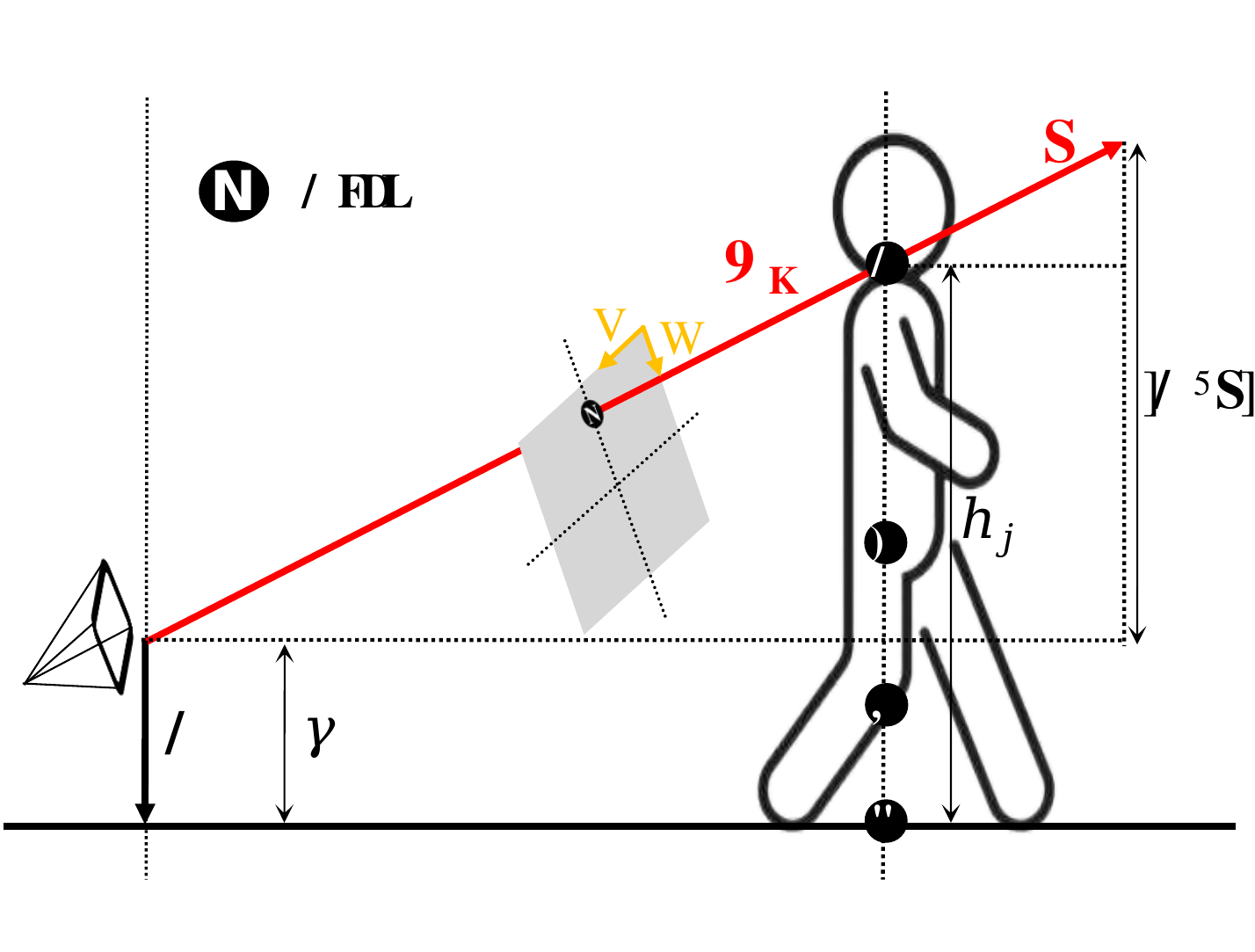}
        \caption{Example of person's location estimation with a neck observation. Any visible joint can be utilized to initialize the person's location. The definition of the parameters and the estimation process are shown in Eq.~\ref{eq-ray-cast}.}
        \label{method-raycasting}
        \vspace*{-0.3in}
\end{figure}

With above detected observation $\mathcal{D}$ in the first frame, through resolving a well-designed over-constrained minimization problem, we construct a \textit{joint-height-based} prior model of the target person, including his position with respect to the camera, which can be used to initialize the Kalman filter tracker. The prior model is represented as:
\begin{equation}
        \mathcal{M}_{J}=\{h_{neck}, h_{hip}, h_{knee}\},
\end{equation}
which corresponds to the heights of the person's neck, hip and knee relative to the ankle, which is assumed to be on the ground at height $0$. Further, we define the person's location as his ankle location, $\mathbf{X}\in\mathbb{R}^3$, in the camera coordinate frame with the ground plane constraint--$\mathbf{N}^T\mathbf{X}+\gamma=0$\cite{ma2004invitation}. This constraint means that the ground plane is defined with a known normal $\mathbf{N}\in\mathbb{R}^3$ at a distance $\gamma>0$ with respect to the optical center (see Fig.~\ref{method-raycasting}).

Based on the above definition, the locations of the other joints can be defined with respect to the ankle location as $\mathbf{X}_j\approx \mathbf{X}+h_j\cdot \mathbf{N}$\cite{fei2021single} where $h_j\in \mathcal{M}_J$. In this work, for model construction, we assume that a full body can be observed in the first frame and define 3D joint positions by the set $\mathcal{X}_{all} = \{\mathbf{X}_{neck},\mathbf{X}_{hip},\mathbf{X}_{knee},\mathbf{X}\}$. We can then obtain the person's location $\mathbf{X}$ and the prior model $\mathcal{M}_J$ by the minimization of the reprojection error:
\begin{equation}
        \{\mathbf{X}, \mathcal{M}_J\}=\mathop{\mathrm{argmin}}_{\mathbf{X}, \mathcal{M}_J} \sum_{i} ||\mathbf{p}_i - g(\mathbf{X}_i)||^2_2,
\label{minimization}
\end{equation}
where $g(\mathbf{X}_i)$ denotes the camera projection function of a 3D point in the camera coordinate frame to the image plane, i.e., $g:\mathbb{R}^3\rightarrow\mathbb{R}^2$. After minimization, the prior model $\mathcal{M}_J$ is obtained, as is the person's location $\mathbf{X}$, which serves as the initial state of the Kalman filter tracker (see next section).

With $\mathcal{M}_J$ constructed, in any of the following frames, the person's location $\mathbf{X}$ can be re-initialized from any one visible joint by Eq.~\ref{eq-ray-cast}\cite{hoiem2008putting} upon target loss and re-identification, without the need to observe the full body. Given the homogeneous pixel coordinates $\bar{\mathbf{p}}_j=[u,v,1]^T$ of a joint, camera intrinsics $\mathbf{K}$, normal vector with $\mathbf{N}$ and its length $\gamma$, and the observed height of a joint $h_j$, we can compute the person's location as:

\begin{equation} \label{eq-ray-cast}
\begin{aligned}
        \mathbf{r}&=\mathbf{K}^{-1}{\bar{\mathbf{p}}_j}\\
        \mathbf{X}_j&=\frac{|\gamma-h_j|}{|\mathbf{N}^T\mathbf{r}|}\cdot\mathbf{r}\\
        \mathbf{X} &\approx \mathbf{X}_j -h_j\cdot\mathbf{N}
\end{aligned}
\end{equation}

Theoretically, we can initialize the person's location by any joint using Eq.~\ref{eq-ray-cast}. However, in practice, because the noise level in the observed joints varies, we prefer to initialize with Eq.~\ref{eq-ray-cast} from the most reliable joint. Empirically, when using AlphaPose\cite{alphaPose}, we observe 1) the upper body is more stable than the lower body for a walking person\cite{bertoni2019monoloco}; 2) the shoulder is more distinguishable than the hip because the shoulder is near the background; and 3) the movement range of the knee is smaller than the movement range of the ankle. Therefore, we use Eq.~\ref{eq-ray-cast} to initialize the person's location with only one joint, in the order of neck, hip, knee, and ankle. 

\subsection{Filtering}  \label{filtering}
After the construction of $\mathcal{M}_J$ and the initialization of the target location, we adopt the unscented Kalman filter (UKF) as our filtering framework due to its advantages in dealing with non-Gaussian noise and non-linear observation function. 
In our UKF-based filtering, the state to be estimated only includes the person's location and velocity on the ground plane, defined as $\mathbf{s}_t=[x_t, y_t, \dot{x}_t,\dot{y}_t]^T$, which is enough to satisfy the need for person following. Further, a constant velocity motion model is assumed, i.e., for motion prediction,
\begin{equation}
        \mathbf{s}_{t+1} = \mathbf{s}_t + \Delta t\cdot [\dot{x}_t,\dot{y}_t,0,0]^T,
\end{equation}
where $\Delta t$ is the time interval between two consecutive frames.
Then we update the state with all visible joints so that each visible joint can contribute a location estimate. In other words, the update step of the Kalman filter depends on the joints that are visible.
As defined in Sec.~\ref{detection}, on the image plane, 2D visible joint positions are a subset of $\{ \mathbf{p}_{neck,t},\mathbf{p}_{hip,t},\mathbf{p}_{knee,t},\mathbf{p}_{ankle,t}\}$. Let $\mathcal{X}_{vis,t}$ be the corresponding 3D joint locations in the camera coordinate frame. Then in the image plane, our observation model used by the UKF is defined by:
\begin{equation}
        \mathbf{z}_t = [g(\mathbf{X}_{j,t})]^T, \mathbf{X}_{j,t}\in \mathcal{X}_{vis,t}.
\label{observation-function}
\end{equation}

\subsection{Data Association}  \label{data-association}

In this filtering framework, we perform data association based on the bounding box information $\mathcal{B}$ instead of the joint information $\mathcal{P}$. 
Because 2D joint detection, compared to the bounding-box detection, is noisy due to occlusion and motion blur, leading to incorrect data association. Therefore, by assuming the person's shape is a cylinder with width $M_W$, we can associate the person's location $\mathbf{X}$ to the bounding box measurement $[u,w]^T$ by calculating the following expected observation:
\begin{equation}
        \bar{u} = g(\mathbf{X})|_x,\quad \bar{w} = f_x\cdot M_W / \mathbf{X}_z,
\label{expected-observation}
\end{equation}
where $\bar{u}$ and $\bar{v}$ represents expected the horizontal component of the bounding-box center and the bounding-box width respectively; $g(\mathbf{X})|_x$ represents the horizontal component of the pixel; $f_x$ is the focal length of the camera. Then our distance metric is defined as:
\begin{equation}
        d = ((\bar{u}-u)^2+(\bar{w}-w)^2)^{1/2},
\end{equation}
where $\bar{u}, \bar{w}$ are from Eq.~\ref{expected-observation} and $u, w$ are from the bounding-box information $\mathcal{B}$. Finally a global nearest neighbor method is utilized to match the measurements to the tracks.

\section{EXPERIMENTS} \label{sec:experiments}
To verify the performance of our proposed \textit{visible-joint-based} method and the whole RPF system, we conduct experiments on different datasets. In this section, we first introduce the datasets, the baselines and implementation details of our experiments. Secondly, the effectiveness of our \textit{visible-joint-based} method is demonstrated by a comparison with the \textit{single-joint-based} and \textit{deep-learning-based} methods on a custom-built dataset. Lastly, we show the superiority of our whole RPF system on a public person following dataset.

\subsection{Datasets}
\begin{figure}[t]
        \centering
        \begin{subfigure}{0.43\linewidth}
                \centering
                \includegraphics[width=\linewidth]{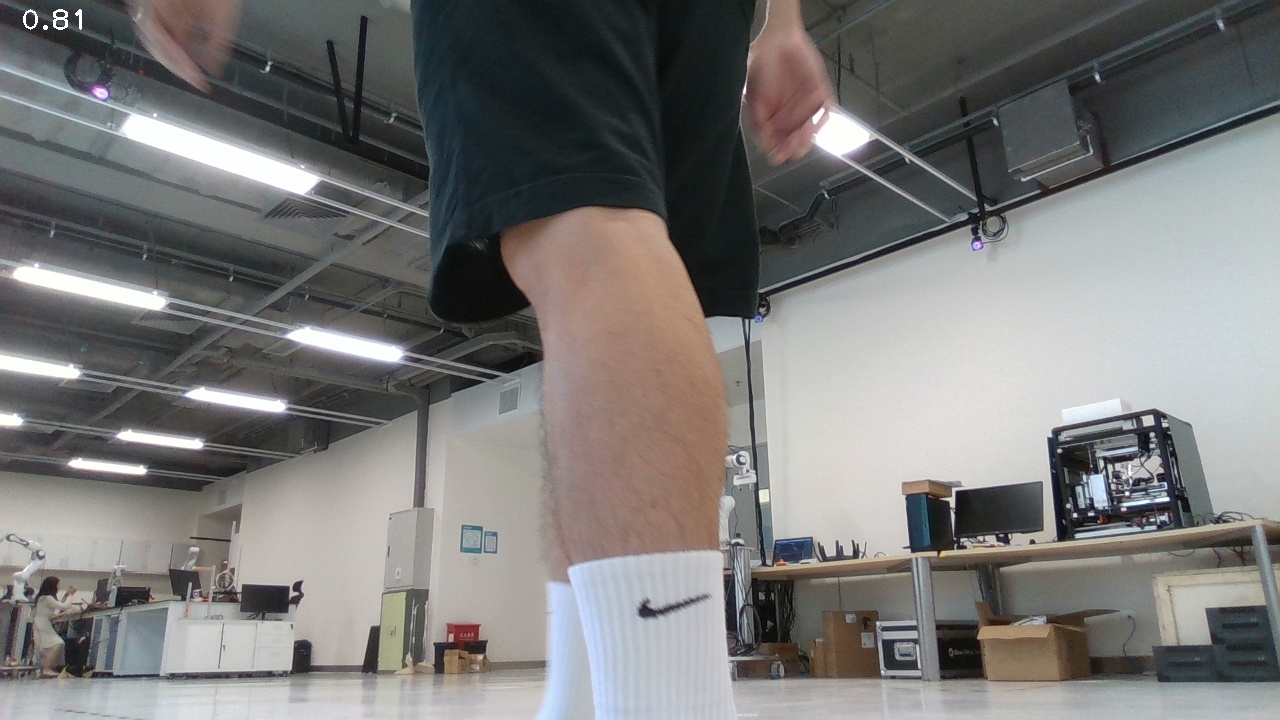}
                \caption{0.5m - 1m}
        \end{subfigure}%
        \hspace{7pt}
        \begin{subfigure}{0.43\linewidth}
                \centering
                \includegraphics[width=\linewidth]{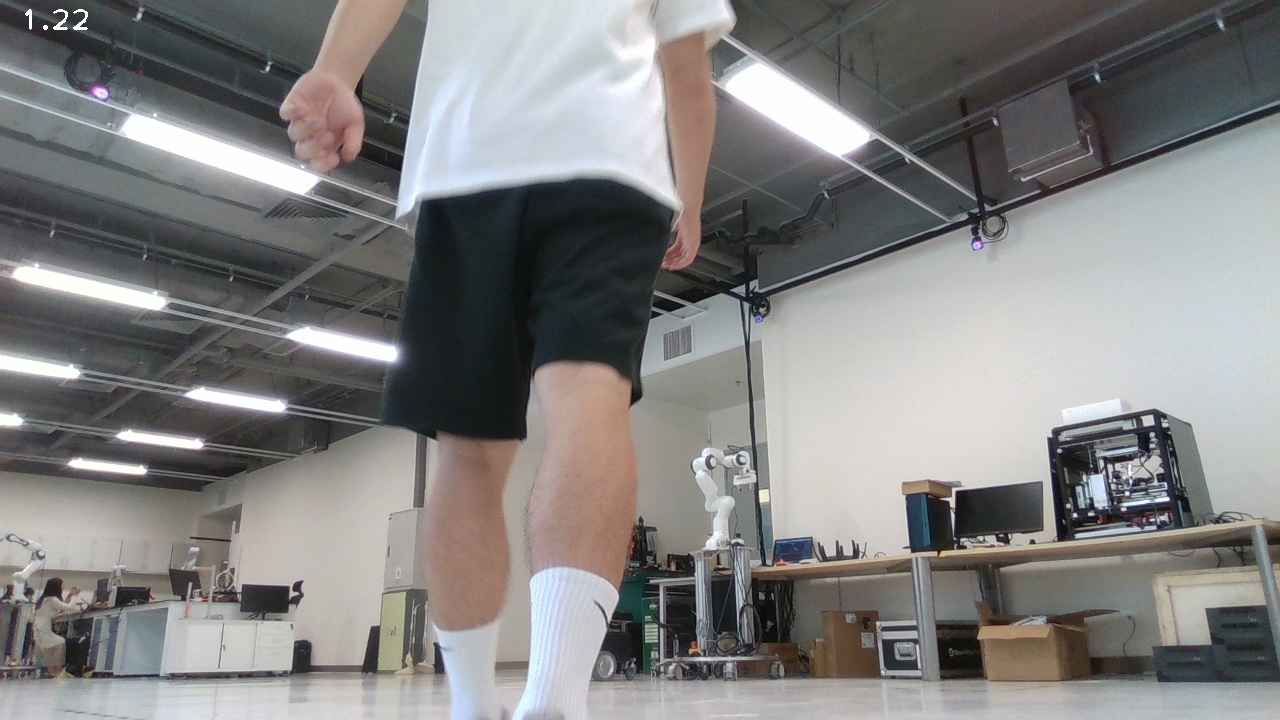}
                \caption{1m - 1.5m}
        \end{subfigure}%
        \vspace{2pt}
        \begin{subfigure}{0.43\linewidth}
                \centering
                \includegraphics[width=\linewidth]{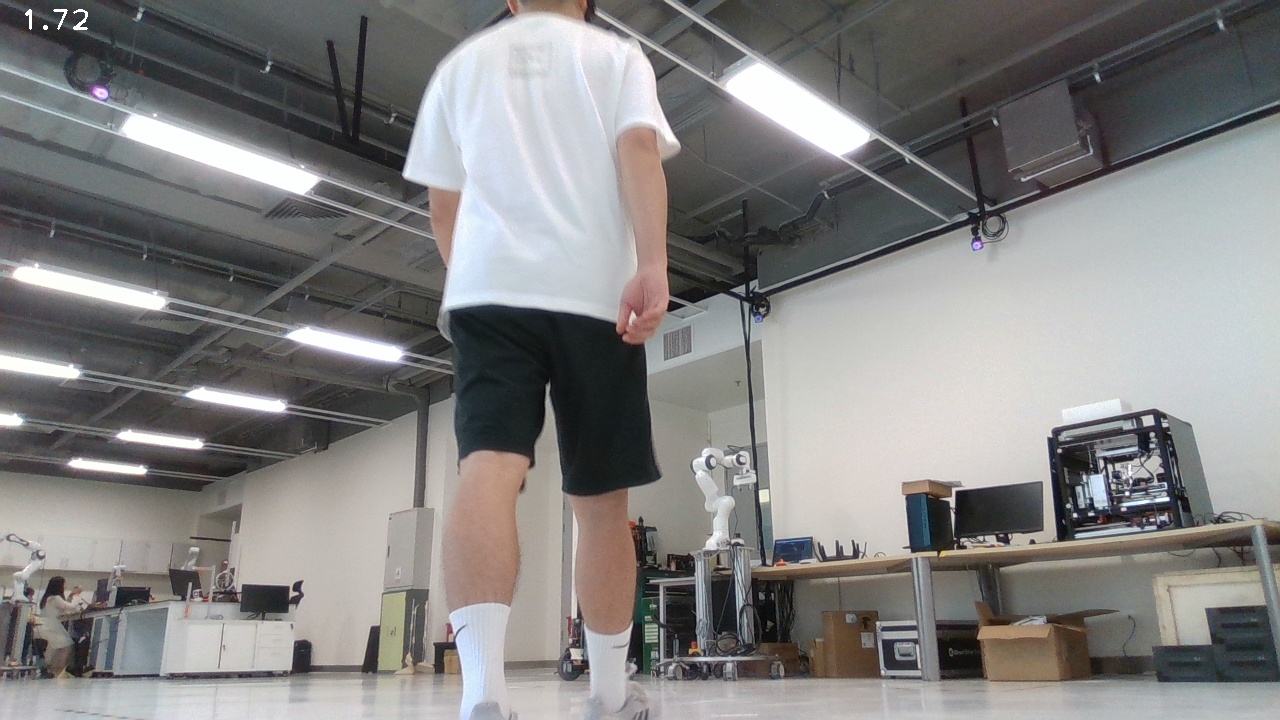}
                \caption{1.5m - 3m}
        \end{subfigure}%
        \hspace{7pt}
        \begin{subfigure}{0.43\linewidth}
                \centering
                \includegraphics[width=\linewidth]{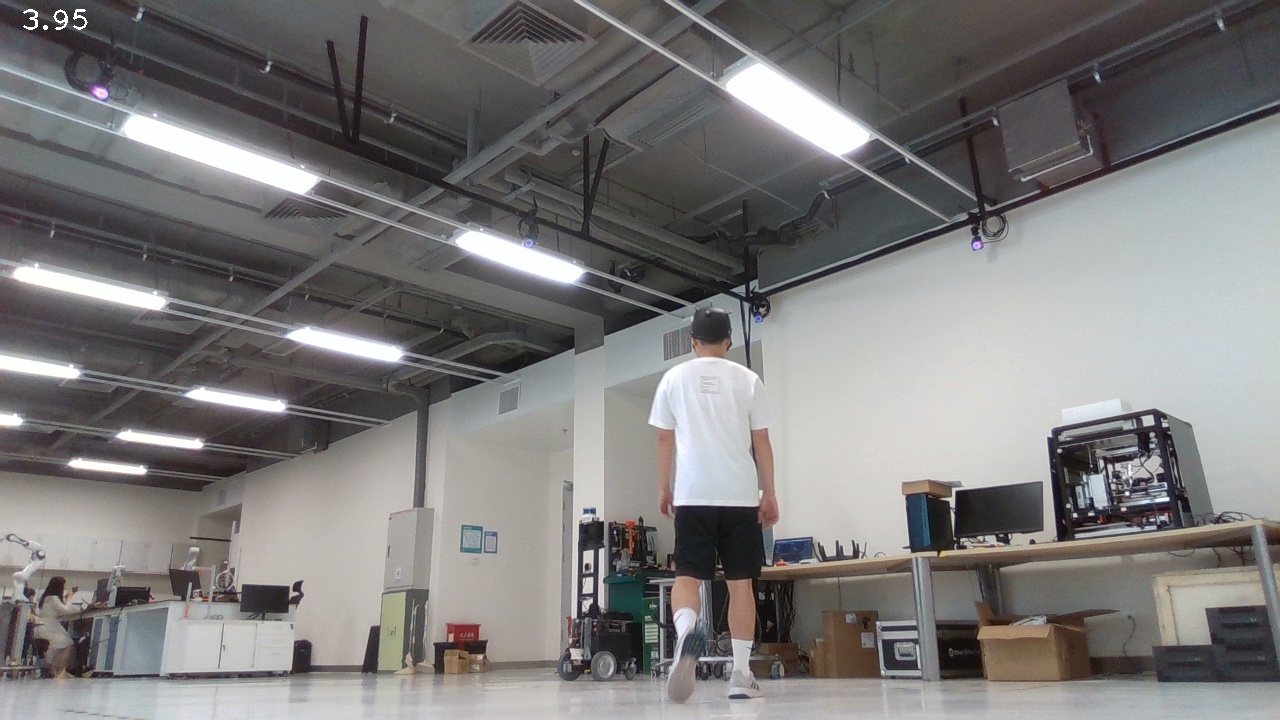}
                \caption{3m - 6m}
        \end{subfigure}%
        \caption{Examples of the custom-built dataset for evaluating the location accuracy of existing methods. From (d) to (a), with the distance decreasing, more joints are occluded. Our method can accurately locate the person under such partial occlusion.}
        \label{dataset}
        \vspace*{-0.15in}
\end{figure}

RPF needs accurate person location estimation and continuous target person tracking. To test these two aspects of our method, we evaluate it on two datasets named as person location (PL) dataset and person tracking (PT) dataset respectively. The PL dataset is a custom-built dataset involving four sequences where the 3D location of a partially occluded person is recorded by a motion capture system or a LiDAR sensor. The distance between the person and the robot varies between 0.5m - 6.0m. Some examples are shown in Fig.~\ref{dataset} where an occluded body is often observed, a difficult condition for location estimation. 
For evaluation of tracking continuity in the 2D image space, the PT dataset, a public dataset\cite{chen2017integrating} with eleven sequences, is used where the target person's ground truth position is represented by a bounding box. This dataset involves challenging situations including illumination change, appearance change and occlusion due to people crossing, which are hard for continuous target person tracking.

\begin{table*}[ht]
        \centering
	\caption{\upshape{Comparison of location performance between our method and esisting baselines on the PL dataset. All sequences are captured within a distance range of 0.5m-2m, except sequence I is within 0.5m-6m. $^\dag$ indicates deep-learning-based methods. \textbf{ALE} (m) represents average location error, \textbf{Recall} is the ratio of the number of recognized frames and that of all frames, and \textbf{WLE} (m) is the proportion of ALE to recall. \XSolidBrush means the algorithm fails to locate the target person in case of failed recognition or the ALE is larger than 5 meters. Our method achieves the lowest WLE in all sequences at 0.11m, 0.09m, 0.10m and 0.08m respectively.}}
	\scalebox{0.9}{
		\begin{tabular}{l|ccc|ccc|ccc|ccc}
			\toprule
                        \multirow{2}*{\textbf{Methods}} 

                        &\multicolumn{3}{c|}{\textbf{I}} &\multicolumn{3}{c|}{\textbf{II}} &\multicolumn{3}{c|}{\textbf{III}} &\multicolumn{3}{c}{\textbf{IV}} \\

			&\textbf{\textit{ALE} $\downarrow$} &\textbf{\textit{Recall} $\uparrow$} &\textbf{\textit{WLE} $\downarrow$}
                        &\textbf{\textit{ALE}} &\textbf{\textit{Recall}} &\textbf{\textit{WLE}}
                        &\textbf{\textit{ALE}} &\textbf{\textit{Recall}} &\textbf{\textit{WLE}}
                        &\textbf{\textit{ALE}} &\textbf{\textit{Recall}} &\textbf{\textit{WLE}} \\
                        
                        \midrule

                        Single-joint-based\cite{koide2020monocular} &\textbf{0.10} &0.64 &0.17 &0.12 &0.04 &3.28 &\XSolidBrush &\XSolidBrush &\XSolidBrush &\XSolidBrush &\XSolidBrush &\XSolidBrush\\

                        MonoLoco$^{\dag}$\cite{monoloc} &1.07 &0.48 &2.22 &\XSolidBrush &\XSolidBrush &\XSolidBrush &\XSolidBrush &\XSolidBrush &\XSolidBrush &\XSolidBrush &\XSolidBrush &\XSolidBrush\\

                        Mono3DBox$^{\dag}$\cite{epropnp} &0.25 &0.66 &0.38 &0.59 &0.21 &2.76 &2.10 &0.09 &22.25 &1.59 &0.39 &4.12 \\

                        Mono3DPose$^{\dag}$\cite{mono3dpose}  &0.36 &0.95 &0.38 &0.51 &\textbf{1.00} &0.51 &0.95 &\textbf{1.00} &0.95 &1.11 &\textbf{1.00} &1.11\\

                        MonoDepth$^{\dag}$\cite{Ranftl2022}  &0.57 &\textbf{1.00} &0.57 &0.32 &\textbf{1.00} &0.32 &0.36 &\textbf{1.00} &0.36 &0.22 &\textbf{1.00} &0.22\\

                        Ours &0.11 &\textbf{1.00} &\textbf{0.11} &\textbf{0.09} &0.98 &\textbf{0.09} &\textbf{0.10} &\textbf{1.00} &\textbf{0.10} &\textbf{0.08} &0.92 &\textbf{0.08}\\                                                
			\bottomrule
	\end{tabular}}
	\label{PL-table}
\end{table*}

\subsection{Baselines}
For evaluating the accuracy of 3D location estimation of our method on the PL dataset, we conduct a comparison with \textit{single-joint-based} and \textit{deep-learning-based} methods introduced in Sec.~\ref{3d-person-location}. All compared methods locate the person in the camera coordinate frame. They are named as follows:
\begin{itemize}
        \item \textbf{Single-joint-based}\cite{koide2020monocular} tracks the person only when the full body is observed including his neck.
        \item \textbf{MonoLoco}\cite{monoloc} could locate the person directly by a neural network with the 2D pose of the person as input.
        \item \textbf{Mono3DBox} predicts the 3D bounding box of the person based on EPro-PnP\cite{epropnp} of which the bottom center is utilized as the estimated location.
        \item \textbf{Mono3DPose} performs person location by predicting his 3D root location through RootNet\cite{mono3dpose}.
        \item \textbf{MonoDepth}, based on MiDas\cite{Ranftl2022}, first obtains the scene depth map and the person's bounding box, then the distance to the person is achieved by averaging the reduced region of the person's depth map. Subsequently, the location of the person can be obtained with the camera reprojection.
\end{itemize}
All estimated locations are transformed to the ground plane and smoothened by the UKF in our evaluation. The main purpose of this comparative experiment is to establish that our solution to location estimation is superior to these SOTA methods in the case of partial occlusion due mostly to the assumption of full-body observation by these SOTA methods during their training.

In addition, we evaluate the person tracking ability of the whole RPF system (involving \textit{visible-joint-based} method and the identification module) as previous RPF works\cite{chen2017integrating}\cite{koide2020monocular}, which regard RPF as a special case of the object tracking. For comparison with object tracking methods, our RPF system, although assuming a calibrated camera to the ground plane to track the person in 3D space, is evaluated in the image space by the associated bounding boxes.
Specifically, we compare our RPF system with popular MOT (multiple object tracking) and SOT baselines, including QDTrack\cite{qdtrack}, Bytetrack\cite{zhang2022bytetrack}, SiamRPN++\cite{li2019siamrpn} and STARK\cite{yan2021learning}. These methods initialize the target person by specifying the ID of the desired person (for MOT) or annotating his bounding box in the first frame (for SOT).

\subsection{Implementation Details}
In our method, bounding-box detection model relies on YOLOX \cite{ge2021yolox} and 2D joint detection on AlphaPose\cite{alphaPose}. The feature extraction model used for appearance description is a deep re-identification model based on CNN\cite{wojke2017simple}. The minimization problem in model construction (Eq.~\ref{minimization}) is solved by the Levenberg-Marquardt algorithm. 

All evaluations are run on a computer with Intel® Core™ i7-10700F CPU @ 2.90GHz and NVIDIA GeForce RTX 2060.
For real robot experiments, a Clearpath Dingo-O and a laptop with Intel(R) Core(TM) i5-10200H CPU @ 2.40GHz and NVIDIA GeForce RTX 1650 are used. A RealSense D435i with $1280\times720$ resolution and 30Hz frequency is mounted on the robot.

\begin{table}[t]
        \centering
	\caption{\upshape{Evaluation of 2D target person tracking between our method and other object tracking baselines on the PT dataset. Our \textit{visible-joint-based} RPF system achieves the best performance with a 97.5\% accuracy.}}
	\scalebox{1.0}{
                \begin{tabular}{lcc}
                        \toprule
                        \textbf{Methods}  &\textbf{Type} &\textbf{Accuracy (\%)}\\
                        \midrule
                        QDTrack\cite{qdtrack} &MOT &48.0\\
                        Bytetrack\cite{zhang2022bytetrack} &MOT &88.6\\
                        \midrule
                        SiamRPN++\cite{li2019siamrpn} &SOT &93.6\\
                        STARK\cite{yan2021learning} &SOT &96.5\\
                        \midrule
                        Single-joint-based &RPF &92.0\\
                        Visible-joint-based &RPF &\textbf{97.5} \\
                        
                        \bottomrule
                \end{tabular}
        }
        \label{PT-accuracy}
        \vspace*{-0.2in}
\end{table}

\subsection{Experimental Results}

\subsubsection{Evaluation of visible-joint-based person location estimation}
This experiment is conducted on the PL dataset, and three metrics are used for evaluation: 1) average location error (ALE) is calculated in the Euclidean space of the ground plane, 2) recall is the ratio of the number of recognized frames to that of all frames, and 3) weighted location error (WLE) is the proportion of ALE to recall, which can evaluate the overall effectiveness of a location estimation method considering both accuracy and robustness. Results are shown in Table~\ref{PL-table}. We can observe that, compared to other methods, our method achieves the lowest WLE in all sequences at 0.11m, 0.09m, 0.10m and 0.08m respectively. 

The \textit{single-joint-based} method\cite{koide2020monocular} achieves a 0.10m ALE and a 64.0\% recall in Sequence I where a full body can be observed occasionally. While in other sequences where only a partially occluded body is observed, the \textit{single-joint-based} method always fails to locate the target person as expected. MonoLoco\cite{monoloc} and Mono3DBox\cite{epropnp} also perform poorly with a low recall and a high ALE, especially on Sequence II to IV. Mono3DPose\cite{mono3dpose} and MonoDepth\cite{Ranftl2022} can work on almost all frames with near 100\% recall but at the expense of a higher ALE compared to our method.

\subsubsection{Evaluation of our RPF system}
The evaluation of our \textit{visible-joint-based} RPF system is conducted independently from considering robot control as a special case of object tracking. Thus we compare the RPF system with other popular object tracking baselines on the PT dataset based on the accuracy metric. Accurate tracking is defined by considering a recognized target person's bounding box as a true positive if the distance between estimated and ground-truth centers of the bounding boxes is less than 50 pixels. 

Results are shown in Table~\ref{PT-accuracy}. STARK\cite{yan2021learning} achieves good performance in MOT and SOT baselines with a 96.5\% accuracy, while our method achieves the best accuracy at 97.5\%. This indicates that in person following, our method can track the target person as well as the popular object tracking methods. Our RPF system, despite the power of our person identification module (which is not the focus in this paper), achieves continuous target person tracking due largely to the ability of tracking under partial occlusion. This can be further verified by the comparison of \textit{single-joint-based} and our \textit{visible-joint-based} RPF systems where only the location estimation method is different. We can observe that if the method changes to \textit{single-joint-based}, the accuracy drops to 92.0\%. 

\subsection{Discussion}
As is shown in Table~\ref{PL-table}, the proposed method achieves the best location estimation performance with a 0.10m ALE, a 98\% recall and a 0.10m WLE on average. Compared to the \textit{single-joint-based} method, it can accurately locate the person even under partial occlusion. Such results indicate that: 1) 2D learning-based pose estimator (AlphaPose\cite{alphaPose}) can perform well under partial occlusion; and 2) our model-based method is able to utilize these well-detected joints to locate the partially occluded person.
Compared to the \textit{deep-learning-based} baselines, our method is also superior because the learning-based methods are  limited in terms of generalization, and are therefore sensitive to environmental changes. On the other hand, our hybrid approach is not dependent on labeled 3D training data, and yet outperforms the baselines.
In conclusion, due to the combination of the robust 2D learning-based joint detector and our model-based location estimator, our \textit{visible-joint-based} method is able to locate the person accurately in robot person following, especially under partial occlusion.

From Table~\ref{PT-accuracy}, we can observe that our method achieves the highest accuracy at 97.5\% which is higher than that of the MOT and SOT baselines. Such a result indicates that, in person following, our method is reliable for not only locating the person accurately but also tracking the person as persistently as these well-designed object tracking methods. The accuracy of our RPF system would drop by 5.5\% if the location method changes to a \textit{single-joint-based} one. This is because, on the PT dataset, there are many situations of partial occlusion, caused by people crossing, corner walls and limited FoV. In above cases, the \textit{single-joint-based} method would lose the person while our method can persistently locate the person and correctly handle these situations of partial occlusion.

\section{CONCLUSION} \label{sec:conclusion}
In this paper, for performing robot person following under partial occlusion, we propose a practical \textit{visible-joint-based} method to locate the person with the observation of any of his neck, hip, knee and ankle joints. Our method takes advantage of a prior model of the tracked person and uses one or more of observed joints for locating the person. The key benefit of our method is that it can locate the person accurately and persistently even under partial occlusion. 
Compared to baselines, our RPF system achieves SOTA target person tracking performance across multiple evaluation metrics.





\bibliographystyle{IEEEtran}
\bibliography{ref}

\begin{thebibliography}{10}
\providecommand{\url}[1]{#1}
\csname url@rmstyle\endcsname
\providecommand{\newblock}{\relax}
\providecommand{\bibinfo}[2]{#2}
\providecommand\BIBentrySTDinterwordspacing{\spaceskip=0pt\relax}
\providecommand\BIBentryALTinterwordstretchfactor{4}
\providecommand\BIBentryALTinterwordspacing{\spaceskip=\fontdimen2\font plus
\BIBentryALTinterwordstretchfactor\fontdimen3\font minus
  \fontdimen4\font\relax}
\providecommand\BIBforeignlanguage[2]{{%
\expandafter\ifx\csname l@#1\endcsname\relax
\typeout{** WARNING: IEEEtran.bst: No hyphenation pattern has been}%
\typeout{** loaded for the language `#1'. Using the pattern for}%
\typeout{** the default language instead.}%
\else
\language=\csname l@#1\endcsname
\fi
#2}}

\bibitem{islam2019person}
M.~J. Islam, J.~Hong, and J.~Sattar, ``Person-following by autonomous robots: A
  categorical overview,'' \emph{The International Journal of Robotics
  Research}, vol.~38, no.~14, pp. 1581--1618, 2019.

\bibitem{goodrich2008human}
M.~A. Goodrich and A.~C. Schultz, \emph{Human-robot interaction: a
  survey}.\hskip 1em plus 0.5em minus 0.4em\relax Now Publishers Inc, 2008.

\bibitem{leigh2015person}
A.~Leigh, J.~Pineau, N.~Olmedo, and H.~Zhang, ``Person tracking and following
  with 2d laser scanners,'' in \emph{2015 IEEE international conference on
  robotics and automation (ICRA)}.\hskip 1em plus 0.5em minus 0.4em\relax IEEE,
  2015, pp. 726--733.

\bibitem{sung2015hierarchical}
Y.~Sung and W.~Chung, ``Hierarchical sample-based joint probabilistic data
  association filter for following human legs using a mobile robot in a
  cluttered environment,'' \emph{IEEE Transactions on Human-Machine Systems},
  vol.~46, no.~3, pp. 340--349, 2015.

\bibitem{yuan2018laser}
J.~Yuan, S.~Zhang, Q.~Sun, G.~Liu, and J.~Cai, ``Laser-based intersection-aware
  human following with a mobile robot in indoor environments,'' \emph{IEEE
  Transactions on Systems, Man, and Cybernetics: Systems}, vol.~51, no.~1, pp.
  354--369, 2018.

\bibitem{wang2017real}
M.~Wang, D.~Su, L.~Shi, Y.~Liu, and J.~V. Miro, ``Real-time 3d human tracking
  for mobile robots with multisensors,'' in \emph{2017 IEEE International
  Conference on Robotics and Automation (ICRA)}.\hskip 1em plus 0.5em minus
  0.4em\relax IEEE, 2017, pp. 5081--5087.

\bibitem{koide2016identification}
K.~Koide and J.~Miura, ``Identification of a specific person using color,
  height, and gait features for a person following robot,'' \emph{Robotics and
  Autonomous Systems}, vol.~84, pp. 76--87, 2016.

\bibitem{linder2016multi}
T.~Linder, S.~Breuers, B.~Leibe, and K.~O. Arras, ``On multi-modal people
  tracking from mobile platforms in very crowded and dynamic environments,'' in
  \emph{2016 IEEE international conference on robotics and automation
  (ICRA)}.\hskip 1em plus 0.5em minus 0.4em\relax IEEE, 2016, pp. 5512--5519.

\bibitem{zhang2019vision}
M.~Zhang, X.~Liu, D.~Xu, Z.~Cao, and J.~Yu, ``Vision-based target-following
  guider for mobile robot,'' \emph{IEEE Transactions on Industrial
  Electronics}, vol.~66, no.~12, pp. 9360--9371, 2019.

\bibitem{zheng2021improving}
L.~Zheng, M.~Tang, Y.~Chen, G.~Zhu, J.~Wang, and H.~Lu, ``Improving multiple
  object tracking with single object tracking,'' in \emph{Proceedings of the
  IEEE/CVF Conference on Computer Vision and Pattern Recognition}, 2021, pp.
  2453--2462.

\bibitem{li2019siamrpn++}
B.~Li, W.~Wu, Q.~Wang, F.~Zhang, J.~Xing, and J.~Yan, ``Siamrpn++: Evolution of
  siamese visual tracking with very deep networks,'' in \emph{Proceedings of
  the IEEE/CVF Conference on Computer Vision and Pattern Recognition}, 2019,
  pp. 4282--4291.

\bibitem{danelljan2019atom}
M.~Danelljan, G.~Bhat, F.~S. Khan, and M.~Felsberg, ``Atom: Accurate tracking
  by overlap maximization,'' in \emph{Proceedings of the IEEE/CVF Conference on
  Computer Vision and Pattern Recognition}, 2019, pp. 4660--4669.

\bibitem{koide2020monocular}
K.~Koide, J.~Miura, and E.~Menegatti, ``Monocular person tracking and
  identification with on-line deep feature selection for person following
  robots,'' \emph{Robotics and Autonomous Systems}, vol. 124, p. 103348, 2020.

\bibitem{monodepth2}
C.~Godard, O.~{Mac Aodha}, M.~Firman, and G.~J. Brostow, ``Digging into
  self-supervised monocular depth prediction,'' in \emph{The International
  Conference on Computer Vision (ICCV)}, October 2019.

\bibitem{park2021dd3d}
D.~Park, R.~Ambrus, V.~Guizilini, J.~Li, and A.~Gaidon, ``Is pseudo-lidar
  needed for monocular 3d object detection?'' in \emph{IEEE/CVF International
  Conference on Computer Vision (ICCV)}, 2021.

\bibitem{openPose}
Z.~{Cao}, G.~{Hidalgo Martinez}, T.~{Simon}, S.~{Wei}, and Y.~A. {Sheikh},
  ``Openpose: Realtime multi-person 2d pose estimation using part affinity
  fields,'' \emph{IEEE Transactions on Pattern Analysis and Machine
  Intelligence}, 2019.

\bibitem{alphaPose}
H.-S. Fang, S.~Xie, Y.-W. Tai, and C.~Lu, ``Rmpe: Regional multi-person pose
  estimation,'' in \emph{2017 IEEE International Conference on Computer Vision
  (ICCV)}, 2017, pp. 2353--2362.

\bibitem{cheng2019occlusion}
Y.~Cheng, B.~Yang, B.~Wang, W.~Yan, and R.~T. Tan, ``Occlusion-aware networks
  for 3d human pose estimation in video,'' in \emph{Proceedings of the IEEE/CVF
  international conference on computer vision}, 2019, pp. 723--732.

\bibitem{bogo2016keep}
F.~Bogo, A.~Kanazawa, C.~Lassner, P.~Gehler, J.~Romero, and M.~J. Black, ``Keep
  it smpl: Automatic estimation of 3d human pose and shape from a single
  image,'' in \emph{European conference on computer vision}.\hskip 1em plus
  0.5em minus 0.4em\relax Springer, 2016, pp. 561--578.

\bibitem{nguyen2022templates}
V.~N. Nguyen, Y.~Hu, Y.~Xiao, M.~Salzmann, and V.~Lepetit, ``Templates for 3d
  object pose estimation revisited: Generalization to new objects and
  robustness to occlusions,'' in \emph{Proceedings of the IEEE/CVF Conference
  on Computer Vision and Pattern Recognition}, 2022, pp. 6771--6780.

\bibitem{hoiem2008putting}
D.~Hoiem, A.~A. Efros, and M.~Hebert, ``Putting objects in perspective,''
  \emph{International Journal of Computer Vision}, vol.~80, no.~1, pp. 3--15,
  2008.

\bibitem{choi2010multiple}
W.~Choi and S.~Savarese, ``Multiple target tracking in world coordinate with
  single, minimally calibrated camera,'' in \emph{European Conference on
  Computer Vision}.\hskip 1em plus 0.5em minus 0.4em\relax Springer, 2010, pp.
  553--567.

\bibitem{liu2020deep}
L.~Liu, W.~Ouyang, X.~Wang, P.~Fieguth, J.~Chen, X.~Liu, and
  M.~Pietik{\"a}inen, ``Deep learning for generic object detection: A survey,''
  \emph{International journal of computer vision}, vol. 128, pp. 261--318,
  2020.

\bibitem{toshev2014deeppose}
A.~Toshev and C.~Szegedy, ``Deeppose: Human pose estimation via deep neural
  networks,'' in \emph{Proceedings of the IEEE conference on computer vision
  and pattern recognition}, 2014, pp. 1653--1660.

\bibitem{girshick2015fast}
R.~Girshick, ``Fast r-cnn,'' in \emph{Proceedings of the IEEE international
  conference on computer vision}, 2015, pp. 1440--1448.

\bibitem{ge2021yolox}
Z.~Ge, S.~Liu, F.~Wang, Z.~Li, and J.~Sun, ``Yolox: Exceeding yolo series in
  2021,'' \emph{arXiv preprint arXiv:2107.08430}, 2021.

\bibitem{sun2019deep}
K.~Sun, B.~Xiao, D.~Liu, and J.~Wang, ``Deep high-resolution representation
  learning for human pose estimation,'' in \emph{Proceedings of the IEEE/CVF
  conference on computer vision and pattern recognition}, 2019, pp. 5693--5703.

\bibitem{newell2017associative}
A.~Newell, Z.~Huang, and J.~Deng, ``Associative embedding: End-to-end learning
  for joint detection and grouping,'' \emph{Advances in neural information
  processing systems}, vol.~30, 2017.

\bibitem{CAO}
Z.~Cao, T.~Simon, S.-E. Wei, and Y.~Sheikh, ``Realtime multi-person 2d pose
  estimation using part affinity fields,'' in \emph{Proceedings of the IEEE
  Conference on Computer Vision and Pattern Recognition (CVPR)}, July 2017.

\bibitem{monoloc}
B.~Lorenzo, K.~Sven, and A.~Alexandre, ``Perceiving humans: From monocular 3d
  localization to social distancing,'' \emph{IEEE Transactions on Intelligent
  Transportation Systems}, vol.~23, no.~7, pp. 7401--7418, 2022.

\bibitem{epropnp}
H.~Chen, P.~Wang, F.~Wang, W.~Tian, L.~Xiong, and H.~Li, ``Epro-pnp:
  Generalized end-to-end probabilistic perspective-n-points for monocular
  object pose estimation,'' in \emph{IEEE Conference on Computer Vision and
  Pattern Recognition (CVPR)}, 2022.

\bibitem{mono3dpose}
G.~Moon, J.~Y. Chang, and K.~M. Lee, ``Camera distance-aware top-down approach
  for 3d multi-person pose estimation from a single rgb image,'' in \emph{2019
  IEEE/CVF International Conference on Computer Vision (ICCV)}, 2019, pp.
  10\,132--10\,141.

\bibitem{ma2004invitation}
Y.~Ma, S.~Soatto, J.~Ko{\v{s}}eck{\'a}, and S.~Sastry, \emph{An invitation to
  3-d vision: from images to geometric models}.\hskip 1em plus 0.5em minus
  0.4em\relax Springer, 2004, vol.~26.

\bibitem{fei2021single}
X.~Fei, H.~Wang, L.~L. Cheong, X.~Zeng, M.~Wang, and J.~Tighe, ``Single view
  physical distance estimation using human pose,'' in \emph{Proceedings of the
  IEEE/CVF International Conference on Computer Vision}, 2021, pp.
  12\,406--12\,416.

\bibitem{bertoni2019monoloco}
L.~Bertoni, S.~Kreiss, and A.~Alahi, ``Monoloco: Monocular 3d pedestrian
  localization and uncertainty estimation,'' in \emph{Proceedings of the
  IEEE/CVF International Conference on Computer Vision}, 2019, pp. 6861--6871.

\bibitem{chen2017integrating}
B.~X. Chen, R.~Sahdev, and J.~K. Tsotsos, ``Integrating stereo vision with a
  cnn tracker for a person-following robot,'' in \emph{International Conference
  on Computer Vision Systems}.\hskip 1em plus 0.5em minus 0.4em\relax Springer,
  2017, pp. 300--313.

\bibitem{Ranftl2022}
R.~Ranftl, K.~Lasinger, D.~Hafner, K.~Schindler, and V.~Koltun, ``Towards
  robust monocular depth estimation: Mixing datasets for zero-shot
  cross-dataset transfer,'' \emph{IEEE Transactions on Pattern Analysis and
  Machine Intelligence}, vol.~44, no.~3, 2022.

\bibitem{qdtrack}
J.~Pang, L.~Qiu, X.~Li, H.~Chen, Q.~Li, T.~Darrell, and F.~Yu, ``Quasi-dense
  similarity learning for multiple object tracking,'' in \emph{IEEE/CVF
  Conference on Computer Vision and Pattern Recognition}, June 2021.

\bibitem{zhang2022bytetrack}
Y.~Zhang, P.~Sun, Y.~Jiang, D.~Yu, F.~Weng, Z.~Yuan, P.~Luo, W.~Liu, and
  X.~Wang, ``Bytetrack: Multi-object tracking by associating every detection
  box,'' in \emph{Proceedings of the European Conference on Computer Vision
  (ECCV)}, 2022.

\bibitem{li2019siamrpn}
B.~Li, W.~Wu, Q.~Wang, F.~Zhang, J.~Xing, and J.~Yan, ``Siamrpn++: Evolution of
  siamese visual tracking with very deep networks,'' in \emph{Proceedings of
  the IEEE Conference on Computer Vision and Pattern Recognition}, 2019, pp.
  4282--4291.

\bibitem{yan2021learning}
B.~Yan, H.~Peng, J.~Fu, D.~Wang, and H.~Lu, ``Learning spatio-temporal
  transformer for visual tracking,'' in \emph{Proceedings of the IEEE/CVF
  International Conference on Computer Vision}, 2021, pp. 10\,448--10\,457.

\bibitem{wojke2017simple}
N.~Wojke, A.~Bewley, and D.~Paulus, ``Simple online and realtime tracking with
  a deep association metric,'' in \emph{2017 IEEE international conference on
  image processing (ICIP)}.\hskip 1em plus 0.5em minus 0.4em\relax IEEE, 2017,
  pp. 3645--3649.

\end{thebibliography}
\end{document}